\documentclass{ecai}
\usepackage{times}
\usepackage{graphicx}
\usepackage{latexsym}

\usepackage{tikz}
\usetikzlibrary{arrows}
\usepackage{float}
\usepackage{amsmath}
\usepackage{amssymb}
\usepackage{amsthm}
\usepackage{stackengine}
\usepackage[linesnumbered,ruled,vlined]{algorithm2e}
\usepackage{pifont}
\newcommand{\cmark}{\ding{51}}
\newcommand{\xmark}{\ding{55}}

\newcommand{\MI}{\textsf{MI}}

\newcommand{\Free}{\textsf{Free}}

\newcommand{\noFacts}{\setminus \mathcal{F}}

\newcommand{\kb}{\mathcal{K}}

\newcommand{\rb}{\mathcal{B}}
\newcommand{\F}{\mathcal{F}}
\newcommand{\R}{\mathcal{R}}
\newcommand{\atoms}{\ensuremath{\mathcal{A}}}
\newcommand{\lang}{\ensuremath{\mathcal{L}}}

\newcommand{\allrbs}{\ensuremath{\mathbb{B}}}

\newcommand{\inc}{\ensuremath{\mathcal{I}}}
\newcommand{\incrb}{\ensuremath{\mathcal{I}^{RB}}}

\newcommand{\posRealInf}{\ensuremath{\mathbb{R}^{\infty}_{\geq 0}}}

\newcommand{\incdrastic}{\ensuremath{\mathcal{I}_{d}}}
\newcommand{\incmi}{\ensuremath{\mathcal{I}_{\mathsf{MI}}}}

\newcommand{\incc}{\ensuremath{\mathcal{I}_{c}}}

\newcommand{\incp}{\ensuremath{\mathcal{I}_{p}}}

\newcommand{\incdrasticrb}{\ensuremath{\mathcal{I}^{RB}_{d}}}
\newcommand{\incmirb}{\ensuremath{\mathcal{I}^{RB}_{\mathsf{MI}}}}

\newcommand{\inccrb}{\ensuremath{\mathcal{I}^{RB}_{c}}}

\newcommand{\incprb}{\ensuremath{\mathcal{I}^{RB}_{p}}}

\theoremstyle{definition}
\newtheorem{definition}[theorem]{Definition}
\newtheorem{proposition}[theorem]{Proposition}
\newtheorem{example}[theorem]{Example}

\ecaisubmission   

\begin{document}

\title{Towards Inconsistency Measurement\\in Business Rule Bases}

\author{Carl Corea \and Matthias Thimm \institute{University of Koblenz-Landau,
Germany, email: \{ccorea,thimm\}@uni-koblenz.de} }

\maketitle
\bibliographystyle{ecai}

\begin{abstract}
  	We investigate the application of inconsistency measures to the problem of analysing business rule bases. Due to some intricacies of the domain of business rule bases, a straightforward application is not feasible. We therefore develop some new rationality postulates for this setting as well as adapt and modify existing inconsistency measures. We further adapt the notion of \emph{inconsistency values} (or \emph{culpability measures}) for this setting and give a comprehensive feasibility study.
\end{abstract}

\section{Introduction}\label{sec:introduction}
\emph{Business rules} have gained major attention in the context of business process management and compliance management \cite{sadiq:2015managing,van:2016BPM,hashmi2018we}. Here, business rules are used to encode policies and laws as a declarative business logic, aimed to ensure that company activities comply with such regulatory controls. For example, the company should only conduct activities as constrained by the set of business rules. Otherwise, the behavior might violate legal regulations, which could result in sensitive fines, or criminal prosecution.
Using business rules to verify the compliance of company activities comes with increased demands on the quality of the business rules themselves. However, as company rule bases are usually maintained by multiple modelers, and in an incremental manner, modeling errors can occur frequently \cite{sadiq:2015managing,batoulis:2017industry,kw2017verification}. For instance, a recent case study with a large insurance company revealed that 27\% of analysed rules contained modeling errors \cite{batoulis:2017industry}. 
Hence, the maintenance of business rule bases is recognized as an important challenge for companies \cite{sadiq:2015managing,corea:2018industry,kw2017verification,nelson2014business} .

A potential problem here is that of \emph{inconsistency}, i.\,e., rules that contradict each other. For example, consider the following business rule base $\rb_1$ (we will formalize syntax and semantics later)
\begin{align*}
    \rb_1 &= \{ \mathit{platinumCustomer}, \mathit{mentalCondition},\\
		&\qquad \mathit{platinumCustomer} \rightarrow \mathit{creditWorthy},\\
		&\qquad \mathit{mentalCondition} \rightarrow \neg \mathit{creditWorthy}\quad \}
\end{align*}
with the intuitive meaning that we have a (platinum) customer who has a mental condition and two general rules stating that 1.) platinum customers are credit worthy, and 2.) a customer with a mental condition is not credit worthy. $\rb_1$ is inconsistent in the classic logical sense, as it entails the contradictory conclusions $creditWorthy$ and $\neg creditWorthy$. Therefore, this rule base cannot be used to draw meaningful conclusions or to correctly regulate process execution. 

To counteract such problems, companies need to be supported with means for the detection and analysis of inconsistencies in business rules, such that experts can resolve inconsistencies. The field of \emph{inconsistency measurement} \cite{Grant:2018,Thimm:2019d} is about analysing inconsistency in logic-based knowledge representations and therefore represents a good candidate for this use-case. In general, an inconsistency measure $\inc$ is a function that assigns a non-negative real value to knowledge base $\kb$, quantifying the inconsistency in $\kb$ with the informal meaning that a higher value reflects a higher degree of inconsistency.


Applying existing inconsistency measures to business rule bases seems straightforward, however, we can identify a conceptional mismatch. 
In the classical setting of inconsistency measurement---that of propositional logic---, knowledge bases are constituted of propositional formulas, where these formulas do not have a distinguishable level of granularity. On the other hand, business rule bases distinguish between \emph{facts} and \emph{rules}. That is, facts have a different conceptual quality as their veracity is unconditionally assumed \cite{graham:2007business}. Thus, a rule base consists of a set of (indisputable) facts and rules.
However, assuming facts as indisputable has strong implications for applying results from inconsistency measurement to this use-case. For example, reconsider the above rule base $\rb_1$. As mentioned, this rule base is inconsistent, but we can see that by removing the fact $mentalCondition$ it becomes consistent.  However, the facts $mentalCondition$, and $platinumCustomer$ are provided by a given case input and have to be kept as-is, even in the scope of inconsistency handling. For instance, one cannot change the mental condition of a customer just to make the set of business rules consistent. Consequently, methods are needed to analyze inconsistency based on a distinction between facts and rules, such that companies can be supported in re-modeling the business rules. 


In this work, we develop means for this use-case as follows:
\begin{enumerate}
    \item We first investigate inconsistency measures for business rule bases in Section 3. To this aim, we propose new postulates that specify expected behavior of inconsistency measures in the business rule base use-case. We show that existing means do not satisfy these requirements and consequently propose new adaptations.
    \item Then, in Section 4, we investigate element-based inconsistency measures, which are useful to pin-point problematic elements in the context of inconsistency handling. Again, we show that existing means are not suitable for our use-case and propose adaptations for a plausible application.
\end{enumerate}
Preliminaries are presented in Section 2. Also, we provide an application example for our proposed means in Section 5, and conclude in Section 6. The proofs of the technical results can be found in the appendix A.

\section{Preliminaries}
To formalise business rule bases, we rely on a simple (monotonic) logic programming language, cf.\ \cite{Gelfond:1991}. For that, we consider a finite set $\atoms$ of atoms. Let $\lang$ be the corresponding set of literals, i.\,e., atoms and negations of atoms. We abbreviate $\overline{\neg a}= a$ and $\overline{a}=\neg a$ for an atom $a$. A \emph{(business) rule base} $\rb$ is then a set of rules of the form
\begin{align}\label{eq:rule}
r\,:\quad  l_{1},\ldots,l_{m} \rightarrow l_{0}.
\end{align}
with $l_{0},\ldots,l_{m}\in\lang$. Let $\allrbs$ be the set of all such rule bases. We abbreviate $head(r)=l_{0}$ and $body(r)=\{l_{1},\ldots,l_{m} \}$. If $body(r)=\emptyset$ we call $r$ a \emph{fact} and simply write $l_{0}$ instead of $\rightarrow l_{0}$. For a rule base $\rb$ let $\F(\rb)\subseteq\rb$ denote the facts in $\rb$ and $\R(\rb)\subseteq\rb$ denote the rules in $\rb$. 

\begin{example}
We recall $\rb_1$ from Section~\ref{sec:introduction}. Then we have
\begin{align*}
	\F(\rb_1) & = \{mentalCondition, platinumCustomer\}\\
	\R(\rb_1) & = \{platinumCustomer \rightarrow creditWorthy,\\
	& \qquad mentalCondition \rightarrow \neg creditWorthy\}.
\end{align*}
\end{example}

A set $M$ of literals is \emph{closed} wrt.\ $\rb$ if for every rule of the form \ref{eq:rule}, if $l_{1},\ldots,l_{m}\in M$ then $l_{0}\in M$. The \emph{minimal model} $M$ of a rule base $\rb$, denoted by $\min(P)$ is the smallest (wrt.\ set inclusion) closed set of literals. A set $M$ of literals is called consistent if it does not contain both $a$ and $\neg a$ for an atom $a$. A program $P$ is called consistent if its minimal model is consistent. 

An inconsistency measure \cite{Grant:2018,Thimm:2019d} is a function $\inc: \allrbs \rightarrow \posRealInf$, where the semantics of the value are defined such that a higher value reflects a higher degree, or severity, of inconsistency.
As the concept of a ''severity`` of inconsistency is not easily characterisable, numerous inconsistency measures have been proposed, see e.g. \cite{Thimm:2018} for an overview. 
A baseline is the drastic inconsistency measure $\incdrastic$ \cite{hunter2008measuring}, which only differentiates between inconsistent and consistent knowledge bases.

In general, all other measures can be divided into formula-centric measures, and atom-centric measures \cite{hunter2010measure}\footnote{We acknowledge there are hybrid forms and some outliers (cf. the discussion in \cite{besnard2016forgetting}), but limit our discussion to these two main perspectives due to space limitations}. In this work, we therefore consider measures representative of these two groups, namely the $\MI-$inconsistency measure \cite{hunter2008measuring}, the problematic inconsistency measure \cite{grant2011measuring} and the contension measure \cite{grant2011measuring}, as shown in Figure \ref{fig:defmeasures}. 

\begin{figure}[H]
	\begin{center}
	\fbox{\begin{minipage}{7.1cm}\small
		\begin{align*}
		    \incdrastic(\kb) & = \left\{\begin{array}{cc}
									1 & \text{if~} \kb\models\perp \\
									0  & \text{otherwise}
									\end{array}\right.\\[0.6ex]
			\incmi(\kb) & =|\MI(\kb)|\\[0.6ex]
			\incp(\kb)	& = |\bigcup_{M\in\MI(\kb)}M|\\[0.6ex]
			\incc(\kb)&= \min\{|\upsilon^{-1}(b)\cap \atoms|\mid\upsilon\models^{3} \kb\}\\[0.6ex]
		\end{align*}
		\end{minipage}}
		\caption{Definitions of the considered inconsistency measures.}
		\label{fig:defmeasures}
	\end{center}
\end{figure}
Formula-centric measures take into account the (number of) formulas responsible for the overall inconsistency. A central approach to measure inconsistency here is derived from minimal inconsistent subsets. Let a rule base $\rb$, the minimal inconsistent subsets $\MI$ of $\rb$ are defined via 
\begin{align*}
	\MI(\rb) = \{M \subseteq \rb \mid M \models \perp, \forall M' \subset M : M' \not\models \perp \}. 
\end{align*}

\begin{example}We recall $\rb_1$. Then we have
\begin{align*}
    \MI(\rb_1) &= \{M_1\}\\
	M_{1}	&= \{platinumCustomer,\\
	        & \qquad platinumCustomer \rightarrow contractuallyCapable,\\
			& \qquad mentalCondition,\\
			& \qquad mentalCondition \rightarrow \neg contractuallyCapable\}
\end{align*}
\end{example}
The $\MI-$inconsistency measure $\incmi$ counts the number of minimal inconsistent subsets. A similar version is the problematic inconsistency measure $\incp$ \cite{grant2011measuring}, which counts the number of distinct formulas appearing in any inconsistent subset.

Atom-centric measures take into account the propositional variables involved in the overall inconsistency. The contension measure $\incc$ quantifies inconsistency by utilizing three-valued interpretations. Here, a three-valued interpretation is a function $v: \atoms \rightarrow \{b,t,f\}$, which assigns every atom to either $b, f$ or $t$, where $t$ and $f$ correspond to the classic logical \textsc{true} and \textsc{false}, and $b$ denotes that there exist conflicting truth values. 
Assuming the \emph{truth order} $\prec_T$ with $f\prec_T b \prec_T t$, the function $v$ is extended to arbitrary formulas as follows: $v(\alpha\wedge\beta) = \min_{\prec_T}(v(\alpha),v(\beta))$, $v(\alpha\wedge\beta) = \max_{\prec_T}(v(\alpha),v(\beta))$, $v(\neg \alpha)=t$ if $v(\alpha)=f$, $v(\neg \alpha)=f$ if $v(\alpha)=t$, and $v(\neg \alpha)=b$ if $v(\alpha)=b$.
We say an interpretation $v$ satisfies a formula $\alpha$, if $v(\alpha) \models t$ or  $v(\alpha) \models b$, which we denote with $v \models^3 \alpha$. Then the contension measure quantifies inconsistency by seeking an interpretation that assigns $b$ to a minimal number of propositions.

\begin{example}
Considering again $\rb_1$, we see that
\begin{align*}
    \incdrastic(\rb_1) = 1 && \incmi(\rb_1) = 1 && \incp(\rb_1) = 4 && \incc(\rb_1) = 1.
\end{align*}
\end{example}
For all considered measures, we see that the measures inherently do not distinguish between facts and rules. Considering our use-case where facts have a different assumption of veracity than rules, this might impede a plausible application. We consequently investigate inconsistency measurement with a distinction between indisputable facts and rules. 




\section{Measures of Inconsistency with indisputable facts}\label{sec:distinctionFactsVsRules}

Consider the following exemplary rule bases $\rb_2$, $\rb_3$, $\rb_4$, defined via
\begin{align*}
	\rb_{2} & = \{a, \neg a\}                                \\   
	\rb_{3} & = \{a, a \rightarrow b, a \rightarrow \neg b\} \\  
	\rb_{4} & = \{a, a \rightarrow b, a \rightarrow \neg b, c, \neg c\}.
\end{align*}
In a business rule management use-case, we are interested only in inconsistencies comprising at least one business rule, as this indicates a human modelling error in the set of business rules. Analysing only such modeling errors is an important basis for re-modelling. In turn, we will not consider inconsistencies such as in $\rb_2$ (this can be handled by existing results from inconsistency measurement), but want to develop new ''rule-based`` inconsistency measures, in the following denoted as $\incrb$, which can specifically assess actual modeling errors, i.\,e., inconsistencies including at least one rule. To this aim, we propose the property of rule-consistency that should be satisfied by rule-based inconsistency measures.
\begin{description}\label{postulate:QC}
	\item[\emph{Rule Consistency} (\textsf{RC})] $\incrb(\rb) = 0$ if and only if for all consistent sets $F'\subseteq \F(\rb)$ , $\R(\rb)\cup F'$ is consistent.
\end{description}

The above rationality postulate is a weakening of the classical postulate \emph{consistency} \cite{thimm:2017compliance}, which requires $\inc(\rb) = 0$ if and only if $\rb$ is consistent. Following \textsf{RC}, measures should assess the degree of inconsistency for $\rb_2$ as 0. Vice versa, in case there is at least one inconsistency which touches at least one rule, the returned inconsistency value should not be 0. 

In any case however, we want to ensure that only conflicts including at least one rule are valued towards the quantification of inconsistency. Atoms appearing only as facts should not alter the degree of inconsistency for rule-based measures, if they do not contradict any rules. We subsequently define the property of fact elision.


\begin{description}\label{postulate:FE}
	\item[\emph{Fact Elision} (\textsf{FE})] If $\forall r \in \rb: head(r) \neq \alpha$ then $\incrb(\rb) = \incrb(\rb\cup\{\overline{\alpha}\})$.
\end{description}

This postulate is closely related to the postulate safe-formula independence (\textsf{SI}) \cite{thimm:2017compliance}, which states that formulas which do not share the signature with the existing propositions of a knowledge base, i.\,e., safe formulas, should not alter the degree of inconsistency. The proposed postulate is a weakening of \textsf{SI}, and states that a formula only needs to be safe w.r.t. the business rules. That is, even if an added formula is \emph{not} safe w.r.t. facts in the knowledge base, this should not alter the score. Consequently, for any $\incrb$, $\incrb(\rb_3\cup c)$ should be equal to $\incrb(\rb_4\cup c)$.

Last, we consider a further aspect of rule-based inconsistency measures. Consider again the rule base $\rb_3$ and the rule base $\rb_5$, defined as
\begin{align*}
	\rb_{3} & = \{a, a \rightarrow b, a \rightarrow \neg b\} &   
	\rb_{5} & = \{a, a \rightarrow b, \neg b\}. 
\end{align*}
In a traditional setting of inconsistency measurement, one could argue that both rule bases are equally inconsistent. 
However, we see that the inconsistency in $\rb_5$ can only be resolved in one way in our setting---namely by modifying or deleting the rule $a \rightarrow b$---as the given facts $a$ and $\neg b$ are indisputable. On the contrary, the inconsistency in $\rb_3$ is caused by contradicting rules
, thus, this inconsistency is more complex to handle and requires attention by domain experts.
To identify such cases, we introduce a third, optional property of rule emphasis. For that, a formula $a \in \rb$ is called a \emph{free formula}, if $a \notin M, \forall M \in \MI(\rb)$. We denote the free formulas of $\rb$ as $\Free(\rb)$.
\begin{description}\label{postulate:RE}
	\item[\emph{Rule Emphasis} (\textsf{RE})] If $B\rightarrow H\notin \rb$ and $B\rightarrow H\notin \Free(\rb\cup\{B\rightarrow H\})$ then $\inc(\rb\cup\{B\rightarrow H\}) > \inc(\rb\cup\{H\})$.
\end{description}
This postulate states that adding a rule to a rule base, where this rule is not a free formula, should increase the inconsistency more than adding only the head of that rule, i.\,e. as a fact. This postulate ensures that measures valuate the conflicts involving contradictory rules as more significant than a conflict resulting from a rule and a non-negotiable fact (as the former type of inconsistency might be more complex to resolve than the latter). 

\begin{example}
For the rule bases $\rb_2$, $\rb_3$, $\rb_4$, and $\rb_5$ from before, we expect a rule-based inconsistency assessment $\incrb$ satisfying the postulates \textsf{RC}, \textsf{FE}, and \textsf{RE} to give
\begin{align*}
    0 = \incrb(\rb_2) &< \incrb(\rb_3) = \incrb(\rb_4) \qquad \text{and}\\
    \incrb(\rb_5) &< \incrb(\rb_3)
\end{align*}
However, for the considered inconsistency measures we get
\begin{align*}
	\incdrastic(\rb_{2}) & =1 & \incdrastic(\rb_{3}) & =1 & \incdrastic(\rb_{4}) & =1 & \incdrastic(\rb_{5}) & =1 \\
	\incmi(\rb_{2})      & =1 & \incmi(\rb_{3})      & =1 & \incmi(\rb_{4})      & =2 & \incmi(\rb_{5})      & =1 \\
	\incp(\rb_{2})       & =2 & \incp(\rb_{3})       & =3 & \incp(\rb_{4})       & =5 & \incp(\rb_{5})       & =3 \\
	%
	%
	%
	%
	%
	\incc(\rb_{2})       & =1 & \incc(\rb_{3})       & =1 & \incc(\rb_{4})       & =2 & \incc(\rb_{5})       & =1
\end{align*}
We see that none of the considered measures is capable of capturing the desired outcome. Specifically, we see that for the above measures (in the following abbreviated as $\inc$ by a slight misuse of notation):
\begin{itemize}
	\item $\inc(\rb_2)>0$ for all measures, thus violating \textsf{RC}.
	\item $\inc(\rb_3)\neq \inc(\rb_4)$ for all measures except $\incdrastic$, thus broadly violating \textsf{FE}.
	\item $\inc(\rb_5) \nless \inc(\rb_3)$ for all measures, thus violating \textsf{RE}.
\end{itemize}
\end{example}
Regarding $\inc(\rb_2) > 0$, this is intuitive, as all considered measures satisfy the postulate of \emph{consistency} (\textsf{CO})\cite{hunter2008measuring}, which demands that the returned value should only be zero iff the rule base is consistent. As a result, we have the following:
\begin{proposition}\label{prop:co}
\textsf{CO} is incompatible with \textsf{RC}.
\end{proposition}

Following from Proposition \ref{prop:co}, virtually all existing inconsistency measures cannot be used as rule-based inconsistency measures, as they uniformly satisfy \textsf{CO} (cf. \cite{thimm:2017compliance}) and thus broadly violate the proposed rationality postulates as motivated from the business use-case. This impedes using existing results in a company context and calls for an adaptation of measures to fit this use-case. In the following, we therefore propose rule-based versions of the original measures. 

\subsection{A baseline for rule-based measures}\label{sec:rbDrastic}

As a baseline measure, the rule-based drastic measure is geared to distinguish between inconsistent and rule-consistent rule bases. To recall, a rule base is not rule-consistent if it contains at least one minimal inconsistent subset, that itself contains at least one rule. To verify this condition, we consider only those minimal inconsistent subsets that do not contain two complementary facts $a, \neg a$.
Formally, define $\MI^{\noFacts}(\rb) = \{M\in \MI(\rb)\mid \neg \exists a\in\atoms: a,\neg a\in M\}$.
If $a,\neg a\in M$ we also call $M$ a pure fact set (note that indeed $a,\neg a\in M$ implies $M=\{a,\neg a\}$).
\begin{example}
We recall $\rb_2$. Then we have
\begin{align*}
    \MI(\rb_2) = \{\{a,\neg a\}\} && \MI^{\noFacts}(\rb_2) = \emptyset
\end{align*}
\end{example}
We are now ready to define a baseline for rule-based measures.
\begin{definition}\label{def:incdrasticrb}
	The rule-based \emph{drastic inconsistency measure} $\incdrasticrb : \allrbs \rightarrow \posRealInf$ is defined as
	\begin{align*}
		\incdrasticrb(\rb) = 
		\begin{cases}
		1 & \text{iff } \MI^{\noFacts}(\rb) \neq \emptyset \\
		0 & \, \text{otherwise}                                                   
		\end{cases}
	\end{align*}
	for $\rb \in \allrbs$.
\end{definition}
In other words, the rule-based drastic measure is 1 if and only if a rule base contains at least one inconsistent subset which is not simply a pair of two complementary facts $a, \neg a$ (and 0 otherwise).

\begin{example}\label{ex:incdrasticrb}
	We recall the business rule bases $\rb_2$ and $\rb_5$, and consider a consistent rule base $\rb_6$
	\begin{align*}
		\rb_{2} & = \{a, \neg a\}                  \\  
		\rb_{5} & = \{a, a \rightarrow b, \neg b\} \\   
		\rb_{6} & = \{a, a \rightarrow b, a\rightarrow c, d\}.
	\end{align*}
	Then, $\incdrasticrb(\rb_5) = 1$, and $\incdrasticrb(\rb_2)=\incdrasticrb(\rb_6)=0$.
\end{example}

\subsection{Rule-Based inconsistency measures based on formulas}\label{sec:rbMinInc}
We continue using the introduced notion of $\MI^{\noFacts}$.

\begin{definition}\label{def:incmirb}
	The rule-based $\MI$-\emph{inconsistency measure} $\incmirb : \allrbs \rightarrow \posRealInf$ is defined as
	\begin{align*}
		\incmirb(\rb) = |\MI^{\noFacts}(\rb)| 
	\end{align*}
	for $\rb \in \allrbs$.    
\end{definition}
Considering only minimal inconsistent subsets without pure fact sets ensures satisfying the requirements of \textsf{RC} and \textsf{FE}, as pure fact $\MI$ are omitted.
\begin{example}\label{ex:incmirb}
	We recall the business rule bases $\rb_2, \rb_3$ and $\rb_5$
	\begin{align*}
		\rb_{2} & = \{a, \neg a\}                                &   
		\rb_{3} & = \{a, a \rightarrow b, a \rightarrow \neg b\} &   
		\rb_{5} & = \{a, a \rightarrow b, \neg b\}.
	\end{align*}
	Then
	\begin{align*}
		\MI(\rb_2) & = \{\{a, \neg a\}\}                                & \MI^{\noFacts}(\rb_2) & = \emptyset                                    \\
		\MI(\rb_3) & = \{\{a, a \rightarrow b, a \rightarrow \neg b\}\} & \MI^{\noFacts}(\rb_3) & = \{\{a, a \rightarrow b, a \rightarrow \neg b\}\} \\
		\MI(\rb_5) & = \{\{a, a \rightarrow b, \neg b\}\}               & \MI^{\noFacts}(\rb_5) & = \{\{a, a \rightarrow b, \neg b\}\}             
	\end{align*}
	and thus $\incmirb(\rb_2)=0$ and $\incmirb(\rb_3)=\incmirb(\rb_5)=1$.
\end{example}


Next, in the original problematic inconsistency measure, the idea is to count the number of formulas contributing in any $\MI$. Thus, in our use-case, an intuitive adaptation is to consider only all problematic rules.

\begin{definition}\label{def:incprb}
	The rule-based \emph{problematic inconsistency measure} $\incprb : \allrbs \rightarrow \posRealInf$ is defined as
	\begin{align*}
		\incprb(\rb)= |\bigcup_{M\in\MI^{\noFacts}(\rb)}M \setminus \F(M)| 
	\end{align*}
	for $\rb \in \allrbs$.    
\end{definition}

\begin{example}\label{ex:problematicII}
	We continue Example \ref{ex:incmirb} and recall
	\begin{align*}
		\MI^1_{\rb_3} & = \{a, a \rightarrow b, a \rightarrow \neg b\} \\
		\MI^1_{\rb_5} & = \{a, a \rightarrow b, \neg b\}               
	\end{align*}
	Then we have
	\begin{align*}
		\MI^1_{\rb_3}\setminus \F(M_1) & = \{a \rightarrow b, a \rightarrow \neg b\} \\
		\MI^1_{\rb_5}\setminus \F(M_1) & = \{a \rightarrow b\}                       
	\end{align*}
	and thus $\incprb(\rb_3) = 2$ and $\incprb(\rb_5) = 1$.
\end{example}

\subsection{Rule-based inconsistency measures based on multi-valued semantics}\label{sec:rbMVSemantics}

 Again, for an adaptation of the contension measure, it is necessary to eliminate conflicts resulting from fact contradictions. Therefore, given a rule base $\rb$ we propose to only consider $\MI^{\noFacts}(\rb)$.

\begin{definition}\label{def:inccrb}
	The rule-based \emph{contension inconsistency measure} $\inccrb : \allrbs \rightarrow \posRealInf$ is defined as
	\begin{align*}
		\inccrb(\rb) = \mathit{min} \{ |v^{-1}(b)\cap \atoms| \mid v \models^3 \bigcup_{M\in\MI^{\noFacts}(\rb)}M \}
	\end{align*}
	for $\rb \in \allrbs$, with $\inccrb(\emptyset) = 0$.    
\end{definition}
\begin{example}\label{ex:inccrb}
	We recall the business rule bases $\rb_2, \rb_3, \rb_4$ and $\rb_5$
	\begin{align*}
		\rb_{2} & = \{a, \neg a\}                                           &   
		\rb_{3} & = \{a, a \rightarrow b, a \rightarrow \neg b\} \\
		\rb_{4} & = \{a, a \rightarrow b, a \rightarrow \neg b, c, \neg c\} &   
		\rb_{5} & = \{a, a \rightarrow b, \neg b\}.
	\end{align*}
	Then
	\begin{align*}
		\bigcup_{M\in\MI^{\noFacts}(\rb_2)}M & = \emptyset                                    \\   
		\bigcup_{M\in\MI^{\noFacts}(\rb_3)}M & = \{a, a \rightarrow b, a \rightarrow \neg b\} \\
		\bigcup_{M\in\MI^{\noFacts}(\rb_4)}M & = \{a, a \rightarrow b, a \rightarrow \neg b\} \\   
		\bigcup_{M\in\MI^{\noFacts}(\rb_5)}M & = \{a, a \rightarrow b, \neg b\}.\\
	\end{align*}
	Then consider $v_1: \{a,b\} \rightarrow \{b,t,f\}$, defined via
	\begin{align*}
	    v_1(a) = t && v_1(b) = b
	\end{align*}
	Then we have $v_1 \models^3 \alpha$ for all formulas $\alpha$ in the considered unions of $\MI^{\noFacts}$. Also, there is no interpretation that assigns $b$ to fewer propositions, and thus $\inccrb(\rb_3) = \inccrb(\rb_4) = \inccrb(\rb_5)=1$ and $\inccrb(\rb_2)=0$. 
\end{example}

\subsection{Analysis}\label{sec:summaryRB}
We now investigate the compliance of the adapted measures with the proposed rationality postulates. We would like to remind the reader that the proofs of the technical results can be found in Appendix A.
\begin{example}
Recall that for the rule bases $\rb_2$, $\rb_3$, $\rb_4$, $\rb_5$ we expect a rule-based inconsistency assessment $\incrb$ satisfying the postulates \textsf{RC}, \textsf{FE}, and \textsf{RE} to give
\begin{align*}
    0 = \incrb(\rb_2) &< \incrb(\rb_3) = \incrb(\rb_4) \qquad \text{and}\\
    \incrb(\rb_5) &< \incrb(\rb_3)
\end{align*}
For our adapted measures we get
\begin{align*}
	\incdrasticrb(\rb_{2}) & =0 & \incdrasticrb(\rb_{3}) & =1 & \incdrasticrb(\rb_{4}) & =1   & \incdrasticrb(\rb_{5}) & =1 \\
	\incmirb(\rb_{2})      & =0 & \incmirb(\rb_{3})      & =1 & \incmirb(\rb_{4})      & =1   & \incmirb(\rb_{5})      & =1 \\
	\incprb(\rb_{2})       & =0 & \incprb(\rb_{3})       & =2 & \incprb(\rb_{4})       & =2   & \incprb(\rb_{5})       & =1 \\
	%
	%
	%
	\inccrb(\rb_{2})       & =0 & \inccrb(\rb_{3})       & =1 & \inccrb(\rb_{4})       & =1   & \inccrb(\rb_{5})       & =1 
	%
\end{align*}
We see that our alterations have improved all measures wrt.\ the rationality postulates, in particular wrt.\  \textsf{RC}.
\end{example}
Table 1 and Table 2 summarize our results regarding the compliance with the rationality postulates motivated by our use case. 
\begin{table}
	\footnotesize
	\parbox{.45\linewidth}{
		\centering
		\begin{tabular}{|l|c|c|c|}
			\hline
			$\inc$      & \textsf{RC} & \textsf{FE} & \textsf{RE} \\
			\hline
			\incdrastic & \xmark      & \xmark      & \xmark      \\
			\incmi      & \xmark      & \xmark      & \xmark      \\
			\incp       & \xmark      & \xmark      & \xmark      \\
			\incc       & \xmark      & \xmark      & \xmark      \\
			\hline
		\end{tabular}
		\caption{Compliance with rationality postulates of the \emph{original} inconsistency measures}
	}
	\hfill
	\parbox{.45\linewidth}{
		\centering
		\begin{tabular}{|l|c|c|c|}
			\hline
			$\incrb$      & \textsf{RC} & \textsf{FE} & \textsf{RE} \\
			\hline
			\incdrasticrb & \cmark      & \cmark      & \xmark      \\
			\incmirb      & \cmark      & \cmark      & \xmark      \\
			\incprb       & \cmark      & \cmark      & \cmark      \\
			\inccrb       & \cmark      & \cmark      & \xmark      \\
			\hline
		\end{tabular}
		\caption{Compliance with rationality postulates of \emph{proposed} rule-based inconsistency measures}
	}
\end{table}

\begin{proposition}\label{prop:compliance1}
$\incdrasticrb,\incmirb,\inccrb$ satisfy \textsf{RC} and \textsf{FE}, and do not satisfy \textsf{RE}.
\end{proposition}
\begin{proposition}\label{prop:compliance2}
$\incprb$ satisfies \textsf{RC}, \textsf{FE} and \textsf{RE}.
\end{proposition}

\section{Inconsistency Values with Indisputable Facts}\label{sec:culpabilityRB}
So far we considered inconsistency measures that assess the \emph{entire} rule base. In the context of inconsistency handling, this is, however, often not sufficient. Companies need to pin-point those formulas in their rule bases that contribute towards the overall inconsistency, e.\,g. as a basis for inconsistency resolution. As a manual analysis of formulas can quickly become unfeasible, the field of inconsistency measurement also studies so-called \emph{inconsistency values}. These are essentially functions which assign a numerical value to individual formulas of a rule base, with the intuition that a higher value indicates a higher blame which a resp.\, formula carries in the context of the overall inconsistency\footnote{Note that measures that determine inconsistency values are also referred to as \emph{culpability measures}.}.

As with inconsistency measures, there have been numerous proposals for specific inconsistency values, see e.\,g. \cite{mcareavey2014computational} for a nice overview. In this work, we consider the Shapley inconsistency value as proposed in \cite{hunter2010measure}, as it is a generalized measure which can be parametrized with arbitrary inconsistency measures. The Shapley inconsistency value uses notions from game theory to determine the blame---also referred to as payoff---that each formula carries w.r.t. the assessment of an arbitrary inconsistency measure. We can consequently directly plug in our proposed rule-based measures. For the following discussion, we assume all rule-based inconsistency measures used to derive inconsistency values satisfy $\textsf{RC}$. Also, we assume the used inconsistency measures satisfy the two basic properties of monotony and free-formula independence \cite{hunter2008measuring}, which are usual desirable properties satisfied by most measures \cite{thimm:2017compliance}.
\begin{description}
	\item[\emph{Monotony} (\textsf{MO})] If $\rb\subseteq \rb'$ then $\inc(\rb)\leq \inc(\rb')$
	\item[\emph{Free-formula independence} (\textsf{IN})] If $\alpha\in\Free(\rb)$ then\\ $\inc(\rb)=\inc(\rb\setminus\{\alpha\})$
\end{description}
\textsf{MO} demands that the addition of information cannot decrease the degree of inconsistency. \textsf{IN} states that free formulas should not affect the degree of inconsistency.

We are now ready to plug rule-based inconsistency measures into the original Shapley inconsistency value.

\begin{definition}
    Let $\incrb$ be a rule-based inconsistency measure, $\rb$ be a rule base and $\alpha \in \rb$. Then, the Shapley inconsistency value of $\alpha$ w.r.t. $\incrb$, denoted $S_\alpha^{\incrb}$ is defined via
    \begin{align*}
        S_\alpha^{\incrb}(\rb) = \sum_{B\subseteq\rb}\frac{(b-1)!(n-b)!}{n!}(\incrb(\rb)-\incrb(\rb\setminus \alpha))
    \end{align*}
    where $b$ is the cardinality of $B$, and $n$ is the cardinality of $\rb$.
\end{definition}
In the following, we consider all elements $\alpha$ of a rule base as a vector ($\alpha_1,\alpha_2,...\alpha_n$), and denote $S^{\incrb}(\rb)$ as the vector of corresponding Shapley inconsistency values, i.\,e., $S^{\incrb}(\rb) = (S_{\alpha_1}^{\incrb}(\rb), S_{\alpha_2}^{\incrb}(\rb),...,S_{\alpha_n}^{\incrb}(\rb) )$. In turn, the Shapley inconsistency value based on the proposed rule-based inconsistency measures satisfies some desirable properties. This result is adapted from \cite{hunter2010measure}, the proofs are analogous.
\begin{proposition}
Let $\incrb$ be a rule-based inconsistency measure, $\rb$ be a rule base and $\alpha \in \rb$. Then, $S^{\incrb}(\rb)$ satisfies:
\begin{itemize}
    \item \textbf{Distribution.} $\sum_{\alpha \in \rb} S_\alpha^{\incrb}(\rb) = \incrb(\rb) $
    \item \textbf{Minimality.} If $\incrb$ satisfies \textsf{IN} and $\alpha$ is a free formula of $\rb$ , then $S_\alpha^{\incrb}(\rb)=0$
\end{itemize}
\end{proposition}

\begin{example}\label{ex:shapleyIgnoringFactContradictions}
Consider the rule base $\rb_7 = \{a,a\rightarrow b,a\rightarrow \neg b,\neg a\}$. Then, for the Shapley inconsistency values w.r.t. $\incdrasticrb$ and $\incmirb$, we have that $S_a^{\incdrasticrb}(\rb) = S_a^{\incmirb}(\rb_7) = \frac{1}{12}+\frac{1}{4} = \frac{1}{3}$. Also, we have that $S_{\neg a}^{\incdrasticrb}(\rb_7) = S_{\neg a}^{\incmirb}(\rb_7) = 0$. Thus, we have $S^{\incdrasticrb}(\rb_7) =(\frac{1}{3},\frac{1}{3},\frac{1}{3},0)$ and $S^{\incmirb}(\rb_7) =(\frac{1}{3},\frac{1}{3},\frac{1}{3},0)$. 
\end{example}
What is nice about Example \ref{ex:shapleyIgnoringFactContradictions} is that it shows that the gist of \textsf{RC} of the rule-based inconsistency measures transfers to the element-based assessment: Only those formulas that are part of at least one $\MI$ which itself contains at least one rule are assigned blame. Still, we see the following problem in using the above Shapley inconsistency value for measuring culpability in our setting. In Example \ref{ex:shapleyIgnoringFactContradictions}, we see that the blame is equally distributed over $\{a,a\rightarrow b,a\rightarrow \neg b\}$ in both assessments. However, facts are viewed as indisputable. This has strong implications for element-based culpability, the simplest one being that facts should not be deleted. In turn, they should also not be assigned with any blame value, as they have to be kept as-is. To capture this requirement, we therefore propose a new property of fact-minimality.
\begin{itemize}
	\item \textbf{Fact-Minimality} $S_{f}^{\incrb}(\rb) = 0$ for any fact $f$ in $\rb$.
\end{itemize}
This is an extension of the minimality property, which is necessary for the intended use-case of viewing facts as indisputable, i.\,e., facts should not be associated with any blame (value) towards the overall inconsistency. As evidenced by Example~\ref{ex:shapleyIgnoringFactContradictions}, the Shapley inconsistency value does not satisfy fact-minimality. Therefore, it is not plausible to apply the Shapley inconsistency value in our use case. We therefore propose an adjusted Shapley inconsistency value. The intuition of our approach is as follows.

The original Shapley inconsistency value essentially assigns responsibilities to a number of formulas (or players) in a coalition. Currently, there is no distinction between facts and rules. Following our use case, the idea is to shift the blame from facts to all rules which are part of the inconsistency for that coalition. We now introduce some notation on this matter for later clarification.

\begin{definition}\label{def:coalPayoff}
    Let $\incrb$ be a rule-based inconsistency measure, $\rb$ be a rule base and $\alpha \in \rb$. Then, the individual Shapley inconsistency coalition value of $\alpha$ w.r.t. $\incrb$ (in a coalition $B \subseteq \rb)$, is defined via
    \begin{align*}
        \mathit{CoalPayoff}_{\alpha,\rb}^{\incrb}(B) = \frac{(b-1)!(n-b)!}{n!}(\incrb(B)-\incrb(B\setminus \alpha))
    \end{align*}
    where $b$ is the cardinality of $B$, and $n$ is the cardinality of $\rb$.
\end{definition}


In our use case, blame should only be assigned to rules. Accordingly, the share of the blame that falls upon facts from any coalition should be shifted away from the facts and equally distributed among the blamable rules, i.\,e., the rules which contribute towards the inconsistency for that coalition. We denote the blame that is shifted from facts to the individual blamable rules, as an \emph{additional payoff}.
\begin{definition}\label{def:additionalPayoff}
Let $\incrb$ be a rule-based inconsistency measure and $\rb$ be a rule base, denote the additional blame for a \emph{rule} $r \in \R(\rb)$ in any coalition $B \subseteq \rb$ as
\begin{align*}
    &\mathit{AddPayoff}_{r,\rb}^{\incrb}(B)\\
    &\qquad = \left\{\begin{array}{ll}
            0& \text{if } r \in \Free(B)\\
            \frac{\sum_{f\in\F(B)}\mathit{CoalPayoff}_{f,\rb}^{\incrb}(B)}{|r\in \R(B) \text{ s.t. } r \notin \Free(B)|} & \text{otherwise}
        \end{array}\right.
\end{align*}
\end{definition}

\begin{example}
   Consider again $\rb_{3} = \{a, a \rightarrow b, a \rightarrow \neg b\}$ and $\rb_{5}  = \{a, a \rightarrow b, \neg b\}$. According to the original Shapley value, the blame is evenly distributed for both rule bases. 
   In $\rb_{3}$, given the premise of indisputable facts, $a$ is not to blame for the overall inconsistency. Rather, the blame value of $a$ should be evenly distributed among $a \rightarrow b$ and $a \rightarrow \neg b$, as both these formulas evenly contribute towards the inconsistency. Next, in $\rb_5$, both $a$ and $\neg b$ are not to blame in our use case. Here, the blame values of $a$ and $\neg b$ should be transferred to $a \rightarrow b$. This is directly in line with the intuition of \textsf{RE}, as for $\rb_5$, the only way to resolve the inconsistency would be to remove $a \rightarrow b$ (thus all blame is relocated to that rule), but for $\rb_3$, one can delete either of the two rules (thus the blame is distributed among both rules). In result, for each coalition, the blame is shifted from facts to the blameable rules via the additional payoff.
\end{example}

We are now ready to define the adjusted Shapley inconsistency value.
\begin{definition}\label{def:adjustedShapley}
    Let $\incrb$ be a rule-based inconsistency measure, $\rb$ be a rule base and $\alpha \in \rb$. Then, the adjusted Shapley inconsistency value of $\alpha$ w.r.t. $\incrb$, denoted $S*_\alpha^{\incrb}$ is defined via
    \footnotesize
    \begin{align*}
        &S*_\alpha^{\incrb}(\rb)\\
        &=
            \begin{cases}
                0 & \text{if } \alpha \in \F(\rb) \\
                \sum\limits_{B\subseteq\rb}( \mathit{CoalPayoff}_{\alpha,\rb}^{\incrb}(B) +  \mathit{AddPayoff}_{\alpha,\rb}^{\incrb}(B) ) & \text{otherwise}
            \end{cases}
    \end{align*}
\end{definition}
The adjusted Shapley value assigns the value of 0 to all facts, and computes the blame value of all rules taking into consideration the additional payoff.

\begin{example}
Consider the rule bases $\rb_{3} = \{a, a \rightarrow b, a \rightarrow \neg b\}$ and $\rb_{5}  = \{a, a \rightarrow b, \neg b\}$. Then for the \emph{adjusted} Shapley inconsistency values w.r.t. $\incdrasticrb$, we have that $S_a^{\incdrasticrb}(\rb_3) = 0, S_{a\rightarrow b}^{\incdrasticrb}(\rb_3) = \frac{1}{3}  (+\frac{1}{3}/2) = \frac{1}{2}$, and $S_{a\rightarrow \neg b}^{\incdrasticrb}(\rb_3) = \frac{1}{3}  (+\frac{1}{3}/2) = \frac{1}{2}$.  Also, we have that $S_a^{\incdrasticrb}(\rb_5) = 0, S_{\neg b}^{\incdrasticrb}(\rb_5) = 0$, and $S_{a\rightarrow \neg b}^{\incdrasticrb}(\rb_3) = \frac{1}{3}  (+\frac{2}{3}) = 1$ . Thus, we have that $S^{\incdrasticrb}(\rb_3) =(0,\frac{1}{2},\frac{1}{2})$ and $S^{\incdrasticrb}(\rb_5) =(0,1,0)$, which is directly in line with \textsf{RE}.
\end{example}

\begin{proposition}\label{prop:adjustedShapley}
The adjusted Shapley value satisfies Distribution, Minimality and Fact-Minimality.
\end{proposition}
 
Proposition \ref{prop:adjustedShapley} shows that our approach follows the same gist as the original Shapley inconsistency value, but shifts the blame from facts to blamable rules as necessary in our use-case.
Next to Distribution and (fact-)Minimality, we can identify a further property for this quantitative assessment $S*_\alpha^{\incrb}(\rb)$.

\begin{proposition}
Let $\incrb$ be a rule-based inconsistency measure, $\rb$ be a rule base and $r \in \R(\rb)$.
\begin{itemize}
    \item \textbf{Rule-Involvement.} If $\incrb$ satisfies \textsf{RC} and $r\in\rb$ is a non-free rule then $S*_{r}^{\incrb}(\rb) > 0$.
\end{itemize}
\end{proposition}
This ensures that rules which are part of any inconsistency have an inconsistency value greater 0.

Next to properties for individual formula assessments, we can also identify properties of the distribution of blame in the vector $S*^{\incrb}$.
\begin{definition}
Let a rule base $\rb$, define $\hat{S}*^{\incrb}(\rb) = max_{\alpha\in\rb} S*_\alpha^{\incrb}(\rb)$.
\end{definition}
\begin{proposition}
Let $\incrb$ be a rule-based inconsistency measure, and $\rb$ be a rule base. Then, following \cite{hunter2010measure}, we have:
\begin{itemize}
    \item \textbf{Rule Consistency'.} $\hat{S}*^{\incrb}(\rb) = 0$ iff $\rb$ is rule consistent.
    \item \textbf{Free formula independence'.}  If $\incrb$ satisfies \textsf{IN} and $\alpha$ is a free formula of $\rb \cup \{\alpha\}$, then $\hat{S}*^{\incrb}(\rb\cup\{\alpha\}) = \hat{S}*^{\incrb}(\rb)$.
    \item \textbf{Upper Bound.} $\hat{S}*^{\incrb}(\rb) \leq \incrb(\rb)$.
    \end{itemize}
\end{proposition}
The first property states that the highest adjusted Shapley inconsistency value can only be 0 in case of a (rule) consistent rule base. The second property states that adding free formulas should not increase any individual values. The last property describes an upper bound.

From the above discussion, we have shown that the proposed adjusted Shapley inconsistency value can be used to assess the distribution of blame for individual rules, relative to the overall inconsistency. Also, the blame (value) will be higher for rules with a higher blame. This information can be useful for companies, e.\,g. for prioritizing which rules to attend to first. To this aim, the elements of a rule base can be ranked by their adjusted Shapley inconsistency value, similar to the approach in \cite{corea:2018industry}.

\begin{definition}[\cite{corea:2018industry}]
Let $\incrb$ be a rule-based inconsistency measure and $\rb$ be a rule base, define the adjusted Shapley culpability ranking over all rules $\alpha \in \rb$, denoted $C(\rb)$ via $\langle \alpha_1,...,\alpha_n\rangle$, s.t. $S*_{\alpha_1}^{\incrb}(\rb) \geq ... \geq S*_{\alpha_n}^{\incrb}(\rb)$.
\end{definition}
Observe that the culpability ranking ranks elements of a rule base by their adjusted Shapley inconsistency values\footnote{As a side effect, this ensures abstraction, i.e. for any rule base $\rb$ and an isomorphism $x$ s.t. $x\rb = \rb', C(x\rb) = C(\rb').$}. As dictated by \emph{Fact-Minimality}, all facts have a value of 0 and rank last (or may be ommitted entirely during inspection). Rather, the ranking represents the blame of all rules in prioritized order and thus provides valuable insights towards inconsistency resolution. In a setting were facts are non-negotiable, it is necessary to adjust the Shapley inconsistency value, as otherwise blame would also be assigned to facts, which could render an undesirable recommendation of deleting a fact.

To show the usefulness of such a culpability ranking and recap the need for the measure adjustments made in this paper, we close with an application example.

\section{Application Example}\label{sec:applicationExample}
Consider the rule base $\rb_{1'}$, defined via
\begin{align*}
	\rb_{1'} = \{ customer, mentalCondition, platinumCustomer,\\
	customer \rightarrow contractuallyCapable,\\
	mentalCondition \rightarrow \neg contractuallyCapable,\\
	mentalCondition \rightarrow \neg platinumCustomer\}
\end{align*}
with the intuitive meaning that we have a (platinum) customer who also has a mental condition, and three rules stating that 1) all customers are generally contractually capable, 2) a person with a mental condition is not contractually capable, and 3) a person with a mental condition is not a platinum customer. We see that $\rb_{1'}$ is inconsistent.

In this Section, we assume a company needs to analyze $\rb_{1'}$ in the scope of inconsistency handling. Furthermore, we assume the facts were provided by a new case input and are thus non-negotiable. 

To begin, $\rb_{1'}$ yields 
\begin{align*}
	\MI(\rb_1') &= \{	M_1, M_{2}\}\\
	M_{1}	&= \{customer,customer \rightarrow contractuallyCapable,\\
			& \qquad mentalCondition,\\
			& \qquad mentalCondition \rightarrow \neg contractuallyCapable\}\\
	M_{2}	&=\{platinumCustomer,mentalCondition,\\
			& \qquad mentalCondition \rightarrow \neg platinumCustomer\}
\end{align*}
Then we have
\begin{align*}
	\incdrastic(\rb_{1'}) & =1 & \incmi(\rb_{1'}) & =2 & \incp(\rb_{1'}) & =6   & \incc(\rb_{1'}) & = 1\\
	\incdrasticrb(\rb_{1'})      & =1 & \incmirb(\rb_{1'})      & =2 & \incprb(\rb_{1'})      & =3   & \inccrb(\rb_{1'})      & = 1
\end{align*}
As can be expected, the rule-based versions of the drastic-, the \MI-, and the contension-measure do not differ from their original counterpart, as we do not have any fact contradictions. However, $\incp$ is highly confusing to modelers in our scenario, as it suggests there are 6 problematic pieces of information. Correctly---w.r.t. the use-case---our adapted $\incprb$ counts 3 problematic pieces of information. Note that the original measures would be even more confusing in our use-case in the presence of  fact contradictions (cf. e.g. the example in Section 3), and would provide only very limited insights towards re-modelling and improving business rules. This becomes even more apparent for inconsistency values:

Assume the company now wants to pin-point elements of the rule base which are responsible for the overall inconsistency as a basis for inconsistency resolution. 

\textbf{Part 1 (Inconsistency handling with existing means).} We recall $\rb_{1'}$. Then we have that
\begin{align*}
    S_{customer}^{\incdrasticrb}(\rb_{1'}) &= \frac{1}{12},\\ S_{mentalCondition}^{\incdrasticrb}(\rb_{1'})&= \frac{3}{12}+\frac{1}{6},\\ S_{platinumCustomer}^{\incdrasticrb}(\rb_{1'}) &= \frac{1}{6},\\
    &etc.
\end{align*}
(Analogously for $\incmirb$). Thus, we have $S^{\incdrasticrb}(\rb_{1'}) =(\frac{1}{12},0.41\overline{6},\frac{1}{6},\frac{1}{12},\frac{1}{12},\frac{1}{6})$ and $S^{\incmirb}(\rb_{1'}) =(\frac{1}{4},0.58\overline{3},\frac{1}{3},\frac{1}{4},\frac{1}{4},\frac{1}{3})$. For both assessments, a recommendation based on the original Shapley value would strongly suggest to delete the fact $mentalCondition$ first. Here, this is not an acceptable recommendation, as one cannot delete the fact that the customer has a mental condition in our setting. Rather, the rules should be deleted or modified. However, even if one would skip the first recommendation based on the original Shapley values, the ranking also does not further distinguish between the remaining facts and rules of the individual $\MI$. We see that the recommendation based on the original Shapley inconsistency value is not plausible and provides very limited value for companies. We will therefore now consider an assessment via our proposed means.

\textbf{Part 2 (Inconsistency handling with the proposed means).} Consider again $\rb_{1'}$. Then we have that 
\begin{align*}
    S*_{mentalCondition}^{\incdrasticrb}(\rb_{1'}) = 0,
\end{align*}
etc. for all facts. Also, we have that 
\begin{align*}
    &S*_{customer\rightarrow contractuallyCapable}^{\incdrasticrb}(\rb_{1'}) &= 0.2\overline{2}\\
    &S*_{mentalCondition\rightarrow \neg contractuallyCapable}^{\incdrasticrb}(\rb_{1'}) &= 0.2\overline{2}\\
    &S*_{mentalCondition \rightarrow \neg platinumCustomer}^{\incdrasticrb}(\rb_{1'}) &= 0.5\overline{5}
\end{align*} (Analogously for $\incmirb$). Thus, we have $S^{\incdrasticrb}(\rb_{1'}) =(0,0,0,0.2\overline{2},0.2\overline{2},0.5\overline{5})$ and $S^{\incmirb}(\rb_{1'}) =(0,0,0,0.5\overline{5},0.5\overline{5},0.8\overline{8})$. A recommendation based on a culpability ranking using these adjusted Shapley values proposes to attend to $mentalCondition\rightarrow\neg platinumCustomer$ first. This makes sense in our example, as this is the only rule in the resp. $\MI$, thus the only option is to delete (or alter) this rule in $M_2$. Then, the recommendation suggests to attend to the remaining two rules with an equal value. This also follows our use-case, as an expert has two possible options in $M_1$. 



\section{Conclusion}

From our discussion, we see that although the field of inconsistency measurement would be a good candidate for supporting companies, a straightforward application is not plausible due to the assumption of non-negotiable facts in the company setting. To this aim, the adapted means presented in this report are a first step towards exploiting the amenities of inconsistency measurement in the scope of business rule bases. Based on recent studies, approaches for inconsistency handling are currently needed from a business perspective and could thus be an interesting application domain for future work \cite{batoulis:2017industry,kw2017verification,sadiq:2015managing,corea:2018industry,diciccio:2017resolving}. 

We would like to point out one specific result from the case-study in \cite{kw2017verification}, namely that companies are not only facing the problem of inconsistent rules, but also the problem of identical (redundant) rules (which could for example result from collaborative modeling or a lack of oversight). As most research in inconsistency measurement is based on sets, applying these results to multi-sets of (business) rules should be further examined, cf. also a recent discussion in \cite{Besnard:2018}.

\medskip

\noindent \textbf{Acknowledgements}: The research reported here was supported by the Deutsche Forschungsgemeinschaft (grant DE 1983/9-1).

\bibliography{ecai}

\newpage
\onecolumn 
\appendix
\section{Proofs of technical results}
\setcounter{theorem}{15}
\begin{proposition}
$\incdrasticrb,\incmirb,\inccrb$ satisfy \textsf{RC} and \textsf{FE}, and do not satisfy \textsf{RE}.
\begin{proof}
For showing \textsf{FE} the following general observation will be useful:
\begin{align}
    \forall r \in \rb: head(r) \neq \alpha \Longrightarrow \MI^{\noFacts}(\rb)=\MI^{\noFacts}(\rb\cup\{\overline{\alpha}\})\label{eq:minofacts}
\end{align}
To see this, let $M\in \MI^{\noFacts}(\rb)$. Then obviously $M\in \MI^{\noFacts}(\rb\cup\{\overline{\alpha}\})$ as well. Let $M\in \MI^{\noFacts}(\rb\cup\{\overline{\alpha}\})$. Then $\overline{\alpha}\notin M$ because 1.) it cannot be that $\alpha\in M$ as this would violate the definition of $\MI^{\noFacts}$ and 2.) there is no rule concluding $\alpha$ by assumption. It follows $M\in\MI^{\noFacts}(\rb)$.

We now consider each measure $\incdrasticrb,\incmirb,\inccrb$ in turn.
\begin{itemize}
    \item 
    We start with the measure $\incdrasticrb$ and \textsf{RC}. Assume $\incdrasticrb(\rb)=0$. Then  $\MI^{\noFacts}(\rb)=\emptyset$ and therefore for all $M\in \MI(\rb)$ there is $a\in\atoms$ with $a,\neg a \in M$. 
    So for all consistent subsets $F'\subseteq \F(\rb)$, $\R(\rb)\cup F'$ is consistent. The other direction is analogous. 
    
    \textsf{FE} follows directly by definition and Equation~(\ref{eq:minofacts}).
    
    For \textsf{RE} consider $\rb=\{a; b; a \rightarrow \neg a\}$. Then $b \rightarrow \neg b\notin \rb$, $b \rightarrow \neg b\notin \Free(\rb\cup\{b \rightarrow \neg b\})$ but $\incdrasticrb(\rb\cup\{\neg b\})=1=\incdrasticrb(\rb\cup\{ b \rightarrow \neg b\})$.
    
    \item We now consider $\incmirb$. The proof of \textsf{RC} is analogous to $\incdrasticrb$. 
    
    \textsf{FE} follows directly by definition and Equation~(\ref{eq:minofacts}).
    
    For \textsf{RE} consider $\rb=\{a; \neg c; b \rightarrow c\}$. Then $a \rightarrow b\notin \rb$, $a \rightarrow b\notin \Free(\rb\cup\{a \rightarrow b\})$ but $\incmirb(\rb\cup\{b\})=1=\incmirb(\rb\cup\{ a \rightarrow b\})$.
    
    \item We now consider $\inccrb$ and  \textsf{RC}. Assume $\inccrb(\rb)=0$. Then there is a three-valued interpretation $v$ with $v(a)=b$ for no $a\in\atoms$ and
    \begin{align*}
       v \models^3 \bigcup_{M\in\MI^{\noFacts}(\rb)}M
    \end{align*}
    Assume $\MI^{\noFacts}(\rb)\neq \emptyset$ and let $M\in \MI^{\noFacts}(\rb)$. Then $v\models^3\alpha$ for each $\alpha\in M$ and as $v^{-1}(b)=\emptyset$, $v(\alpha)=t$. This is a contradiction as $M$ is inconsistent. It follows $\MI^{\noFacts}(\rb)= \emptyset$ and therefore for all consistent sets $F'\subseteq \F(\rb)$ , $\R(\rb)\cup F'$ is consistent. On the other hand, if for all consistent sets $F'\subseteq \F(\rb)$ , $\R(\rb)\cup F'$ is consistent then $\MI^{\noFacts}(\rb)= \emptyset$ and $v$ can be defined via $v(a)=t$ for all $a\in \atoms$ showing $\inccrb(\rb)=0$.
    
    \textsf{FE} follows directly by definition and Equation~(\ref{eq:minofacts}).
    
    For \textsf{RE} consider $\rb=\{a; \neg c; b \rightarrow c\}$. Then $a \rightarrow b\notin \rb$, $a \rightarrow b\notin \Free(\rb\cup\{a \rightarrow b\})$ but $\inccrb(\rb\cup\{b\})=1=\inccrb(\rb\cup\{ a \rightarrow b\})$.
\end{itemize}
\end{proof}
\end{proposition}
\begin{proposition}\label{prop:compliance2}
$\incprb$ satisfies \textsf{RC}, \textsf{FE} and \textsf{RE}.
\begin{proof}
    The postulates \textsf{RC} and \textsf{FE} follow from a similar reasoning as for $\incmirb$ in the proof of Proposition~\ref{prop:compliance2}. For \textsf{RE}, let $B\rightarrow H\notin \rb$ and $B\rightarrow H\notin \Free(\rb\cup\{B\rightarrow H\})$. First observe that $\MI^{\noFacts}(\rb)\subseteq \MI^{\noFacts}(\rb\cup\{H\rightarrow B\})$. Consider then some $M\in \MI^{\noFacts}(\rb\cup\{H\})\setminus \MI^{\noFacts}(\rb)$. It follows $H\in M$. As $B\rightarrow H\notin \Free(\rb\cup\{B\rightarrow H\})$ it follows that there is $X\subseteq \rb$ such that $H\in \min(X\cup\{B\rightarrow H\})$ and the set of facts appearing in $X$ is consistent (otherwise $B\rightarrow H$ could not be part of any $N\in\MI^{\noFacts}(\rb\cup\{B\rightarrow H)\}$). Then $M'=M\setminus\{H\}\cup X\cup\{B\rightarrow H\}$ is inconsistent. Let $M''\subseteq M'$ be minimal inconsistent. It follows $M''\in \MI^{\noFacts}(\rb\cup\{B\rightarrow H)\}$. This means that every $M\in \MI^{\noFacts}(\rb\cup\{H\})\setminus \MI^{\noFacts}(\rb)$ can be transformed to an $M''\in \MI^{\noFacts}(\rb\cup\{B\rightarrow H)\}$ by replacing $H$ with the rule $B\rightarrow H$ and further rules and facts to derive $B$. As this replacement introduces (at least) the rule $B\rightarrow H$ we get $\incprb(\rb\cup\{B\rightarrow H\})>\incprb(\rb\cup\{H\})$.
\end{proof}
\end{proposition}

\setcounter{theorem}{25}
\begin{proposition}\label{prop:adjustedShapley}
The adjusted Shapley value satisfies Distribution, Minimality and Fact-Minimality.
\begin{proof}
We adress the three parts individually.
\begin{itemize}
    \item For showing Distribution, let us rewrite the Definition of the adjusted Shapley inconsistency value as
        \begin{align*}
        S*_\alpha^{\incrb}(\rb)
        =
            \begin{cases}
                0 & \text{if } \alpha \in \F(\rb) \\
                \sum\limits_{B\subseteq\rb} \mathit{CoalPayoff}_{\alpha,\rb}^{\incrb}(B) + 
                \sum\limits_{B\subseteq\rb}\mathit{AddPayoff}_{\alpha,\rb}^{\incrb}(B) & \text{otherwise}
            \end{cases}
    \end{align*}
    We are now interested in the sum of all adjusted Shapley values (for all elements of a rule base $\rb$).
    \begin{align*}
        \sum_{\alpha \in \rb} S*_\alpha^{\incrb}(\rb)
        &=\sum_{\alpha \in \rb}
            \begin{cases}
                0 & \text{if } \alpha \in \F(\rb) \\
                \sum\limits_{B\subseteq\rb} \mathit{CoalPayoff}_{\alpha,\rb}^{\incrb}(B) + 
                \sum\limits_{B\subseteq\rb}\mathit{AddPayoff}_\alpha^{\incrb}(B) & \text{otherwise}
            \end{cases}\\
        &=\sum_{\alpha \in \rb}\sum\limits_{B\subseteq\rb} \mathit{CoalPayoff}_{\alpha,\rb}^{\incrb}(B) - \sum_{f \in \F(\rb)}\sum\limits_{B\subseteq\rb} \mathit{CoalPayoff}_{f,\rb}^{\incrb}(B)+\sum_{\alpha \in \R(\rb)}\sum\limits_{B\subseteq\rb}\mathit{AddPayoff}_{\alpha,\rb}^{\incrb}(B)\\
        \intertext{Following \cite{hunter2010measure}, the first summand can be rewritten.}
        &=\incrb(\rb) - \sum_{f \in \F(\rb)}\sum\limits_{B\subseteq\rb} \mathit{CoalPayoff}_{f,\rb}^{\incrb}(B)+\sum_{\alpha \in \R(\rb)}\sum\limits_{B\subseteq\rb}\mathit{AddPayoff}_{\alpha,\rb}^{\incrb}(B)\\
        \intertext{Then}
        &=\incrb(\rb) - \sum_{f \in \F(\rb)}\sum\limits_{B\subseteq\rb} \mathit{CoalPayoff}_{f,\rb}^{\incrb}(B)+\sum_{\alpha \in \R(\rb)}\sum\limits_{B\subseteq\rb} \left\{\begin{array}{ll}
            0& \text{if } r \in \Free(B)\\
            \frac{\sum_{f\in\F(B)}\mathit{CoalPayoff}_{f,\rb}^{\incrb}(B)}{|r\in \R(B) \text{ s.t. } r \notin \Free(B)|} & \text{otherwise}
            \end{array}\right.  \\
        &=\incrb(\rb) - \sum_{f \in \F(\rb)}\sum\limits_{B\subseteq\rb} \mathit{CoalPayoff}_{f,\rb}^{\incrb}(B)+\sum_{r \in \Free(\R(\rb))}\sum\limits_{B\subseteq\rb} 0 + \sum_{r \not\in \Free(\R(\rb))}\sum\limits_{B\subseteq\rb} \frac{\sum_{f\in\F(B)}\mathit{CoalPayoff}_{f,\rb}^{\incrb}(B)}{|r\in \R(B) \text{ s.t. } r \notin \Free(B)|}\\
        &=\incrb(\rb) - \sum_{f \in \F(\rb)}\sum\limits_{B\subseteq\rb} \mathit{CoalPayoff}_{f,\rb}^{\incrb}(B) + \sum_{r \not\in \Free(\R(\rb))}\sum\limits_{B\subseteq\rb} \frac{\sum_{f\in\F(B)}\mathit{CoalPayoff}_{f,\rb}^{\incrb}(B)}{|r\in \R(B) \text{ s.t. } r \notin \Free(B)|}\\
        &=\incrb(\rb) - \sum_{f \in \F(\rb)}\sum\limits_{B\subseteq\rb} \mathit{CoalPayoff}_{f,\rb}^{\incrb}(B) + \sum\limits_{B\subseteq\rb} \sum_{f\in\F(B)}\mathit{CoalPayoff}_{f,\rb}^{\incrb}(B)\\ 
        &=\incrb(\rb) - \sum_{f \in \F(\rb)}\sum\limits_{B\subseteq\rb} \mathit{CoalPayoff}_{f,\rb}^{\incrb}(B) + \sum_{f\in\F(B)}\sum\limits_{B\subseteq\rb}\mathit{CoalPayoff}_{f,\rb}^{\incrb}(B)\\
        &=\incrb(\rb)
    \end{align*}
    
    \item Minimality follows from a combination of Definitions \ref{def:additionalPayoff}, \ref{def:coalPayoff} and \ref{def:adjustedShapley}. Observe that any element in a rule base $\rb$ is either a fact or a rule, thus we discuss this individually.
    For any fact $f$, $S*_f^{\incrb}(\rb) = 0$ via Definition \ref{def:adjustedShapley}. Thus we have that if a free formula $\alpha$ is a fact, $S*_\alpha^{\incrb}(\rb) = 0$. Continuing with rules, following the free-formula independence requirement, we have that if a rule $r$ is a free-formula, then $\incrb(\rb) = \incrb(\rb\setminus r).$ Then we see that the last part ($\incrb(\rb)-\incrb(\rb\setminus r)$) in Definition \ref{def:coalPayoff} always equals out to zero for free formulas (This follows the original proof in \cite{hunter2010measure}). Last, via Definition \ref{def:additionalPayoff}, we see that any rule which is a free formula automatically is assigned an additional payoff of zero. Concluding, the adjusted shapley value via Defintion \ref{def:adjustedShapley} is zero for rules which a free formulas.
    
    \item Fact-minimality directly follows from Definition \ref{def:adjustedShapley}, i.e., facts are always assigned a value of zero.
    
\end{itemize}
\end{proof}
\end{proposition}

\setcounter{theorem}{26}
\begin{proposition}
Let $\incrb$ be a rule-based inconsistency measure, $\rb$ be a rule base and $r \in \R(\rb)$.
\begin{itemize}
    \item \textbf{Rule-Involvement.} If $\incrb$ satisfies \textsf{RC} and $r\in\rb$ is a non-free rule, then $S*_{r}^{\incrb}(\rb) > 0$.
\end{itemize}
\begin{proof}
A useful observation is that for a non-free formula $\alpha$, all minimal inconsistent subsets that $\alpha$ is contained relate to coalitions, where removing $\alpha$ from that coalition resolves the inconsistency.
Then we see that the last part ($\incrb(\rb)-\incrb(\rb\setminus r)$) in Definition \ref{def:coalPayoff} will be $\neq 0$ for those coalitions. Also, any non-free rule is at least part of one minimal inconsistent subset $M_1$. Thus, to show rule-involvement, we see that the smallest adjusted Shapley value a non-free rule can receive is $((m-1)!(n-m)!)/n!)+\sum_{f\in\F(M_1)}(((m-1)!(n-m)!)/n!)
/(|r\in \R(M_1) \text{ s.t. } r \notin \Free(M_1)|))$, where $m$ is the cardinality of $M_1$, and $n$ is the cardinality of $\rb$. 
\end{proof}
\end{proposition}

\begin{proposition}
Let $\incrb$ be a rule-based inconsistency measure, and $\rb$ be a rule base. Then, following \cite{hunter2010measure}, we have:
\begin{itemize}
    \item \textbf{Rule Consistency'.} $\hat{S}*^{\incrb}(\rb) = 0$ iff $\rb$ is rule consistent.
    \item \textbf{Free formula independence'.}  If $\incrb$ satisfies \textsf{IN} and $\alpha$ is a free formula of $\rb \cup \{\alpha\}$, then $\hat{S}*^{\incrb}(\rb\cup\{\alpha\}) = \hat{S}*^{\incrb}(\rb)$.
    \item \textbf{Upper Bound.} $\hat{S}*^{\incrb}(\rb) \leq \incrb(\rb)$.
    \end{itemize}
    \begin{proof}
    To show Rule Consistency', observe that as dictated by Rule-Involvement, we have that $S*_{r}^{\incrb}(\rb) > 0$ for any non-free rule $r\in\rb$. So, for an inconsistent rule base $\rb'$, we have that $\hat{S}*^{\incrb}(\rb') > 0$. From the other side, a consistent rule base $\rb$ only contains free formulas. Here, following minimality, we see that indeed $\hat{S}*^{\incrb}(\rb')$ would be 0. To show Free formula independence' observe that for free $\alpha$ and $\incrb$ satisfying \textsf{IN} we have $\incrb(B)=\incrb(B\cup\{\alpha\})$. Then for arbitrary $\gamma\in\rb\cup\{\alpha\}$, $B\subseteq \rb\cup\{\alpha\}$, $b=|B|$, and $n=|\rb|$ we have
    \begin{align*}
        \incrb(B)-\incrb(B\setminus\{\gamma\}) = \incrb(B\cup\{\alpha\})-\incrb((B\cup\{\alpha\})\setminus\{\gamma\})
    \end{align*}
    and therefore
    \begin{align*}
        &\mathit{CoalPayoff}_{\gamma,\rb\cup\{\alpha\}}^{\incrb}(B) + \mathit{CoalPayoff}_{\gamma,\rb\cup\{\alpha\}}^{\incrb}(B \cup\{\alpha\}) \\
        =& \frac{(b-1)!(n+1-b)!}{(n+1)!}(\incrb(B)-\incrb(B\setminus\{\gamma\})) + \frac{b!(n-b)!}{(n+1)!}(\incrb(B\cup\{\alpha\})-\incrb((B\cup\{\alpha\})\setminus\{\gamma\}))\\
        =& (\frac{(b-1)!(n+1-b)!}{(n+1)!} + \frac{b!(n-b)!}{(n+1)!} ) (\incrb(B)-\incrb(B\setminus\{\gamma\}))\\
        =& \frac{(b-1)!(n+1-b)! + b!(n-b)!}{(n+1)!}(\incrb(B)-\incrb(B\setminus\{\gamma\}))\\
        =& \frac{(b-1)!(n-b)! (n+1-b+b)}{n! (n+1)}(\incrb(B)-\incrb(B\setminus\{\gamma\}))\\
        =& \frac{(b-1)!(n-b)!}{n!}(\incrb(B)-\incrb(B\setminus\{\gamma\}))\\
        =& \mathit{CoalPayoff}_{\gamma,\rb}^{\incrb}(B)
    \end{align*}
    yielding $S*_\gamma^{\incrb}(\rb)=S*_\gamma^{\incrb}(\rb\cup\{\alpha\})$ for all $\gamma$. Also $S*_\alpha^{\incrb}(\rb\cup\{\alpha\})=0$ and therefore
    \begin{align*}
        \hat{S}*^{\incrb}(\rb\cup\{\alpha\}) & = \max_{\gamma\in\rb\cup\{\alpha\}} S*_\gamma^{\incrb}(\rb\cup\{\alpha\})\\
            & = \max_{\gamma\in\rb} S*_\gamma^{\incrb}(\rb)\\
            & = \hat{S}*^{\incrb}(\rb)
    \end{align*}
    
    Last, Upper Bound follows directly from Distribution, i.e., considering a knowledge base $\rb$ and a rule-based inconsistency measure $\incrb$, we see that the sum of $S*_{\alpha}^{\incrb}(\rb)$ over all $\alpha \in \rb$ equals $\incrb(\rb)$. In turn, for any $\alpha_i \in \rb$, $S*_{\alpha_i}^{\incrb}(\rb)$ cannot be larger than $\incrb(\rb)$.
    \end{proof}
\end{proposition}

\end{document}